\documentclass[letterpaper, 10 pt, conference]{ieeeconf}  

\IEEEoverridecommandlockouts                              

\usepackage{graphics} 
\usepackage{epsfig} 
\usepackage{mathptmx} 
\usepackage{times} 
\usepackage{amsmath} 
\usepackage{amssymb}  
\usepackage{bm}
\usepackage{algorithm}
\usepackage{multirow}
\usepackage{verbatim}
\usepackage{graphicx}
\usepackage{hyperref}

\usepackage{subcaption}
\usepackage{lipsum}
\usepackage{mwe}

\usepackage{titlesec}
\titlespacing{\section}{0pt}{*0.6}{*0.6}
\titlespacing{\subsection}{4pt}{*0.5}{*0.5}

\usepackage{booktabs} 
\title{\LARGE \bf
An Experimental Validation and Comparison of Reaching Motion Models for Unconstrained Handovers: Towards Generating Humanlike Motions for Human-Robot Handovers
\vspace{-4mm}
}

\author{Wesley P. Chan$^{1}$, Tin Tran$^{1}$, Sara Sheikholeslami$^{1,2}$, Elizabeth Croft$^{1}$
\thanks{$^{1}$Faculty of Engineering, Monash University}
\thanks{$^{2}$Mechanical Engineering, University of British Columbia}
\thanks{\textcopyright2021 IEEE. Personal use of this material is permitted.  Permission from IEEE must be obtained for all other uses, in any current or future media, including reprinting/republishing this material for advertising or promotional purposes, creating new collective works, for resale or redistribution to servers or lists, or reuse of any copyrighted component of this work in other works.
}
} 
\begin{document}

\maketitle
\thispagestyle{empty}
\pagestyle{empty}

\begin{abstract}
The Minimum Jerk motion model has long been cited in literature for human point-to-point reaching motions in single-person tasks. While it has been demonstrated that applying minimum-jerk-like trajectories to robot reaching motions in the joint action task of human-robot handovers allows a robot giver to be perceived as more careful, safe, and skilled, it has not been verified whether human reaching motions in handovers follow the Minimum Jerk model. To experimentally test and verify motion models for human reaches in handovers, we examined human reaching motions in unconstrained handovers (where the person is allowed to move their whole body) and fitted against 1) the Minimum Jerk model, 2) its variation, the Decoupled Minimum Jerk model, and 3) the recently proposed Elliptical (Conic) model. Results showed that Conic model fits unconstrained human handover reaching motions best. Furthermore, we discovered that unlike constrained, single-person reaching motions, which have been found to be elliptical, there is a split between elliptical and hyperbolic conic types. We expect our results will help guide generation of more humanlike reaching motions for human-robot handover tasks.

\end{abstract}

\section{INTRODUCTION}
        \label{sec:introduction}

The service robotics sector has been growing rapidly in recent decades with development for applications ranging from homecare \cite{badii2009} to manufacturing \cite{Chan2020}. For humanoid robots working in such applications, \textit{object handover} is a fundamental task that arises frequently (e.g., handing over a tool in a factory, or a TV remote at home); thus, efficient performance of handovers is essential to their effectiveness.

The study of handovers is a diverse and growing area. Researchers comparing handover reaching motions have found that robots employing human-like reaching motions are preferred \cite{Shibata1997,Huber2008}. Eye gaze in handovers has been investigated \cite{Moon2014} with use of human-like gaze behaviours used by robots demonstrated to increase handover efficiency. Studies on dynamics of the physical handover have identified grip force control strategies used by humans for coordinating object transfer \cite{Chan2012}, and demonstrated successful application to human-robot handovers \cite{Chan2013,Parastegari2016}. Human body motion and kinematics have been studied, and classifiers built for predicting handover events \cite{Strabala2013,Pan2017}. 
These and other works have shown that enabling robots to employ humanlike behaviours improves human-robot handovers. However, human behaviours in handovers are not always well understood. Hence, towards improving human-robot handovers, there has been much research on first understanding and modelling human behaviours in the different aspects of handovers.

Herein, we focus on arm motions in the reaching phase of handovers. One widely adopted approach for enabling fluent human-robot handovers is to enable human-like reaching motions for robots \cite{Shibata1997,Huber2008}. Various models have been proposed to describe human handover reaching motions, and applied to human-robot handovers. However, empirical validations of such models to show whether these models fit natural human handover motions are lacking - this has largely remained an assumption. Hence, with the goal of improving human-robot handovers through enabling generation of more humanlike reaching motions, this paper focuses on  validating and comparing proposed models with a dataset of observed natural human handover motions. Having a validated model of human handover motion will allow us to implement more human-like humanoid motions towards enabling more fluent human-robot cooperations.
        
\section{RELATED WORK}
        \label{sec:related_works}
        Human handovers typically happen very fast (within $\sim$1.2s according to our dataset). To enable online generation of handover trajectories, a computationally fast model is required. Although more advanced and complex models such as those drawn from biomechanics or Inverse Optimal Control for describing human motions exist, they are less suitable for the target application of real-time human-robot handovers. In this section, we describe existing models that have been proposed for and/or applied to human-robot handovers.

The \textit{Minimum Jerk} trajectory model for human reaching motions was proposed in the 1980s \cite{Hogan1982}. Since then, this model has been widely accepted and applied to robot reaching motions in various human-robot interaction tasks, including handovers, and continues to be used for modeling human handover motions in recent works \cite{Li2015,Landi2019}. Early on, velocity profiles resembling the Minimum Jerk trajectory were applied to 1D tabletop human-robot handovers \cite{Shibata1997}. Compared to typical industrial robot trajectories, a minimum-jerk-like trajectory allowed the robot to be perceived as more careful, pleasant, and skilled. Later, the Minimum Jerk trajectory was applied to 3D handover tasks seated at a table \cite{Huber2008}. Similarly, a Minimum Jerk trajectory allowed the robot to be perceived as safer with shorter reaction time from human receivers. While the original applications of the Minimum Jerk model assumed that the path, like human reaching motions, is, in general, approximately straight \cite{Morasso1981,Abend1982}, it has later been shown that reaching motions in handovers are curved \cite{Huber2009}. As a result, a \textit{Decoupled Minimum Jerk} trajectory model was proposed, which decouples the 3D reaching motion into two Minimum Jerk trajectories - one in the z direction, and one in the orthogonal xy plane \cite{Huber2009}. This results in a curved trajectory, but still has the characteristic that the ending portion of the trajectory is straight. While existing studies demonstrated benefits of employing human-like motions for human-robot handovers, they have not specifically fitted the proposed models to experimental data to evaluate how well these models describe human handover motions.

Recently, an \textit{Elliptical} (Conic) model has been proposed for human reaching motions. This model has been experimentally shown to accurately fit motions in single-person seated pick-and-place tasks \cite{Sheikholeslami2018}. It was suggested that the model can potentially be applied to human-robot interaction tasks such as handovers. However, it has not been shown that human reaching motions in the joint action task of handover also follow the proposed Elliptical model. 

Literature shows that reaching motions vary with task context \cite{Marteniuk1987}. There is also online coordination and adaptation between giver and receiver actions during handovers \cite{Yamane2013,Chan2012}. Hence, reaching motions in single person tasks and in joint action tasks such as handovers may differ. Furthermore, existing studies have been largely limited to constrained tasks, where participants seated at a table handling one generically shaped object \cite{Shibata1997,Huber2009,Sheikholeslami2018}. This is quite different to every day handovers that service robots will need to perform. Thus, our aim is to first verify if human reaching motions in unconstrained handovers, handing over everyday objects, fit existing proposed models, and evaluate which model fits best. In the following section, we describe and provide mathematical definitions for the models considered.

\section{REACHING MOTION MODEL}
        \label{sec:method}
        \subsection{Elliptical (Conic) Motion Model}
It has been empirically shown that the trajectory of human reaching motions in pick-and-place tasks follows a planar curve, and that an elliptical curve achieves a good fit to the path drawn out by the projection of the trajectory on a plane \cite{Sheikholeslami2018}. The general conic section equation is 
\begin{equation}
\label{eq:conic}
Ax^2 + Bxy + Cy^2 + Dx + Ey + F = 0
\end{equation}

Sheikholeslami et al. \cite{Sheikholeslami2018} describes a procedure for computing the coefficients of Eq \ref{eq:conic}. The best fit plane of the trajectory is first computed, and the trajectory points projected onto this plane. Singular Value Decomposition is then used to compute the coefficients. The fitted conic equation can then be classified through its discriminant as: elliptical if $B^2 - 4AC < 0$ and $A \neq C$, hyperbolic if $B^2 - 4AC > 0$, and parabola if $B^2 - 4AC = 0$.
In \cite{Sheikholeslami2018} it was found that a majority of their observed data were elliptical.

An ellipse can also be expressed in parametric form as
\begin{gather}
\label{eq:ellipse_parametric}
    \begin{bmatrix} x(\theta) \\ y(\theta) \end{bmatrix}
    = 
    \begin{bmatrix} x_c \\ y_c \end{bmatrix}
    +
    \begin{bmatrix} \cos(\tau) & -\sin(\tau) \\ \sin(\tau) & \cos(\tau)  \end{bmatrix}
    \begin{bmatrix} a\cos(\theta) \\ b\sin(\theta)\end{bmatrix}
\end{gather}
where $x_c$, $y_c$ is the ellipse center, $\tau$ the inclination angle, and $a, b$ the semi major and minor axes. Using this form, we can generate a trajectory by specifying time dependence of $\theta$.

\subsection{Minimum Jerk Motion Model}
The Minimum Jerk model hypothesizes that human arm motion between two points minimizes jerk over the entire path \cite{Hogan1982}. Given position over time $r(t)$, jerk is the third derivative, $\dddot{r}(t)$. Path smoothness can be measured by
\begin{equation}
S = \int_{0}^{t_f} \dddot{r}(t)^2 dt
\label{eq:smoothness}
\end{equation}
where $t_f$ is the duration of the motion. The Minimum Jerk trajectory minimizes $S$, and takes on the form:
\begin{equation}
    r(t) = a_0 + a_1t + a_2t^2 + a_3t^3 + a_4t^4 + a_5t^5,
\label{eq:min-jerk}
\end{equation}
where the coefficients $a_i$ are determined by the boundary conditions, i.e., position, velocity, and acceleration at trajectory start and end points.

\subsection{Decoupled Minimum Jerk Motion Model}
Literature has shown that if we decouple the z axis motion from the xy plane motion by specifying two Minimum Jerk trajectories with different durations for the two components, we obtain a curved trajectory that more closely resembles human handover reaching motions \cite{Huber2009}.
This Decoupled Minimum Jerk trajectory is described by
\begin{equation}
    r_z(t) = a_{0z} + a_{1z}t + a_{2z}t^2 + a_{3z}t^3 + a_{4z}t^4 + a_{5z}t^5,
\label{eq:decoupled-min-jerk-z}
\end{equation}
\begin{equation}
    r_{xy}(t) = a_{0xy} + a_{1xy}t + a_{2xy}t^2 + a_{3xy}t^3 + a_{4xy}t^4 + a_{5xy}t^5,
\label{eq:decoupled-min-jerk-xy}
\end{equation}
where $r_z(t)$ is the trajectory in the z direction, with duration $t_z$, and $r_{xy}(t)$ is the trajectory in the xy plane, with duration $t_{xy}$. The coefficients $a_{iz}$ and $a_{ixy}$ are again determined by boundary conditions. The Decoupled Minimum Jerk trajectory results in a curved path in 3D space, residing in a plane orthogonal to the xy plane. The ratio between $t_z$ and $t_{xy}$ is determined through optimization as described in \cite{Huber2009}.   
        
\section{HUMAN HANDOVER DATASET}
        \label{sec:experiment}
\begin{figure}[t]
\vspace{2mm}
\centering
\includegraphics[width=.45\textwidth]{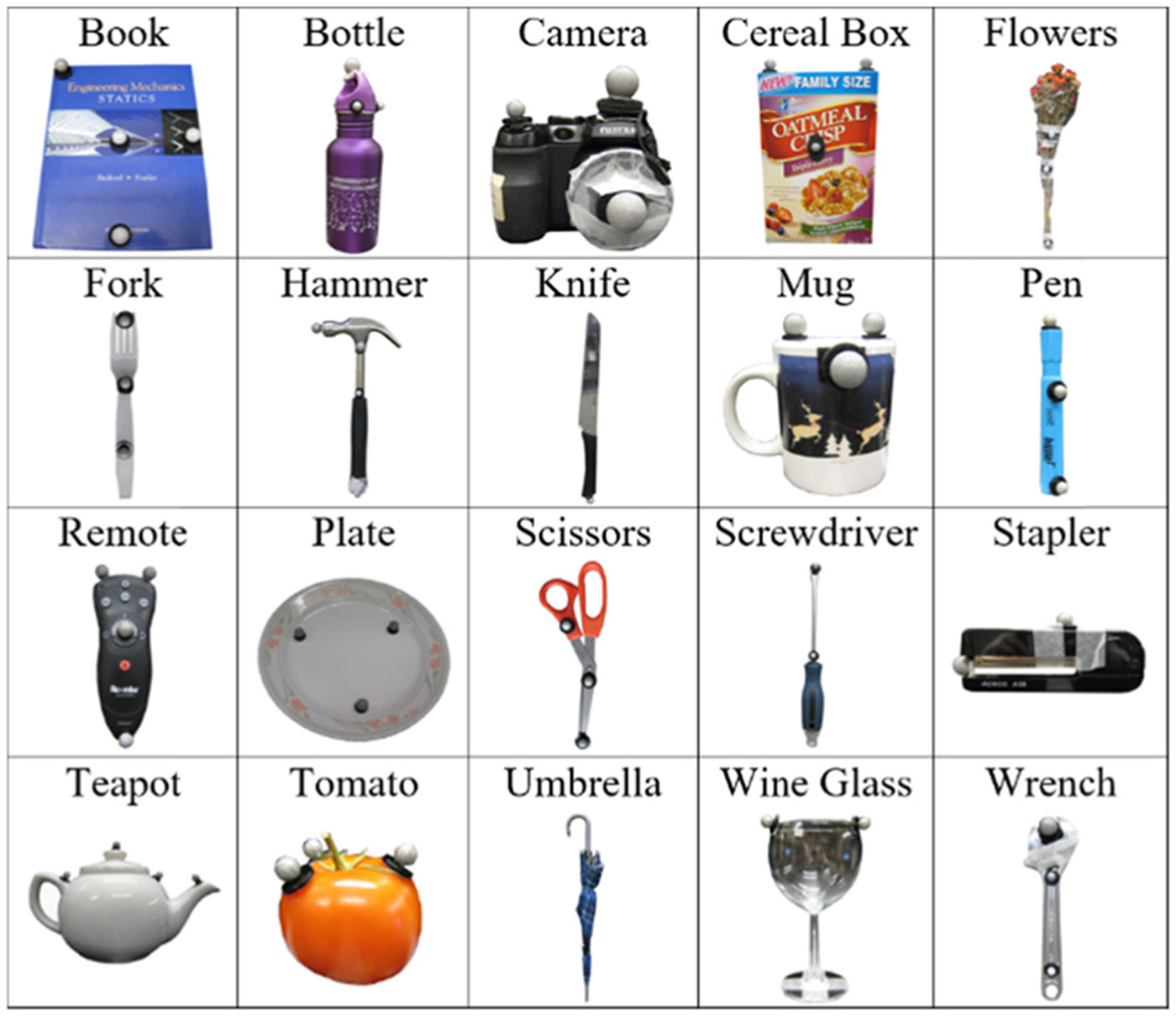}
\caption{The twenty common everyday objects used for handovers in the dataset. (Image from \cite{Chan2020handover}.)
}
\label{fig:objects}
\end{figure}

\begin{figure}[t]
\centering
\includegraphics[width=.40\textwidth]{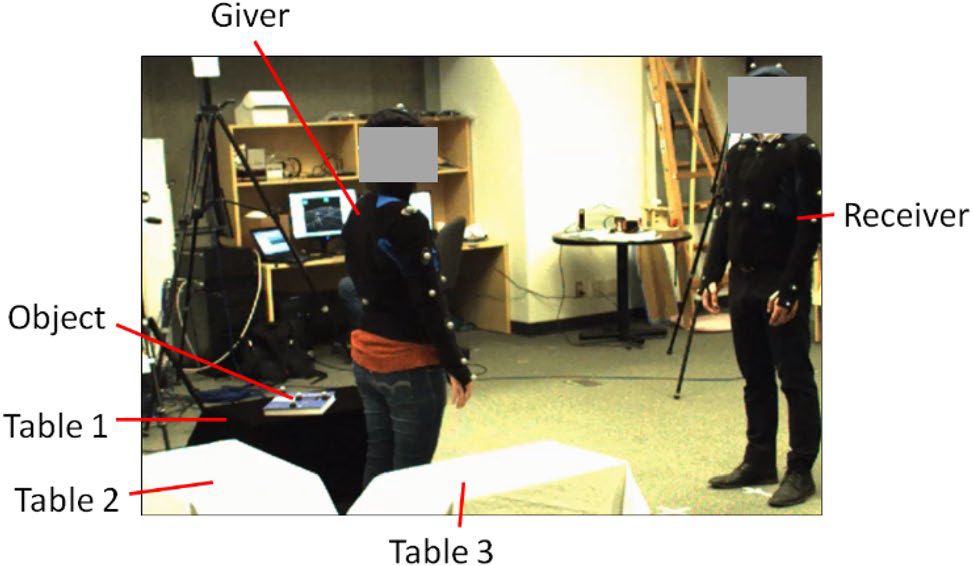}
\caption{Data collection setup for collecting human handover motions in the dataset. (Image from \cite{Chan2020handover}.)
}
\label{fig:setup}
\end{figure}

        We used the publicly available Handover Orientation and Motion Capture Dataset\footnote{\url{https://bridges.monash.edu/articles/Handover_Orientation_and_Motion_Capture_Dataset/8287799}} \cite{Chan2020handover}, which contains 1200 unconstrained human handover trajectories performed by twenty participants recorded by a Vicon motion capture system. In the dataset, twenty common objects (Fig. \ref{fig:objects}) were used, and each handover started with giver and receiver standing facing each other, and one of the objects placed on a table left, right, or behind the giver. Fig. \ref{fig:setup} shows the data collection setup. The giver picked up the object and handed it over to the receiver using unconstrained motion (i.e., they were allowed to move their whole body). Additional details of the data collection procedure can be found in \cite{Chan2020handover}.

\section{ANALYSIS}
        \label{sec:analysis}
        \subsection{Handover Segmentation}
\begin{figure}[t]
    \vspace{2mm}
    \centering
    \includegraphics[width=0.40\textwidth]{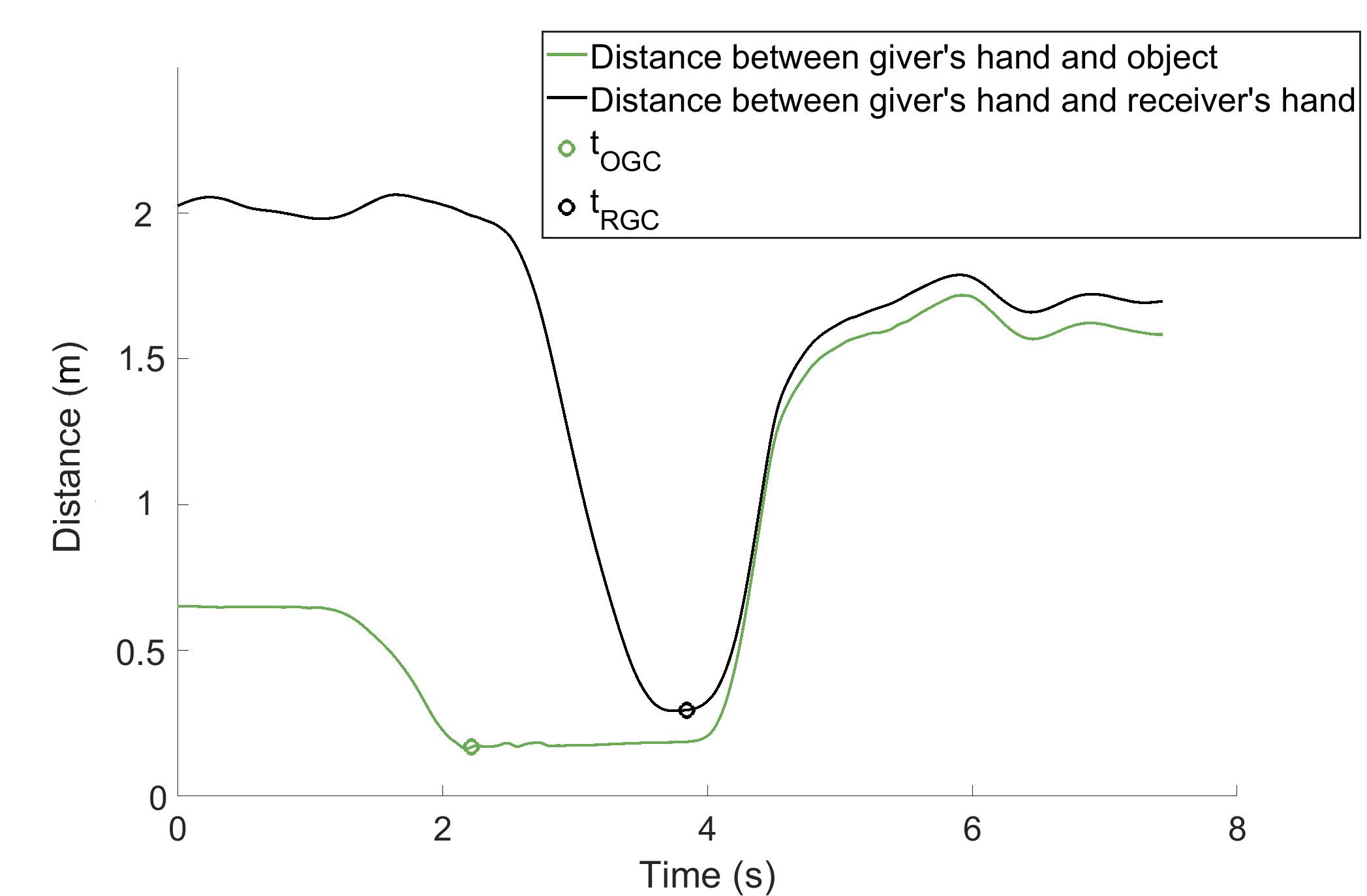}
    \caption{Plot of distance between giver's hand and object, and distance between giver's hand and receiver's hand in an example handover trial.}
    \label{fig:prelim_search}
    \end{figure}

Spectral analysis of the dataset reveals most signal power reside below 7Hz, so we filtered all handover trajectories using a 10 Hz low pass Butterworth filter. To segment the handover reaching motion, we first identified the instance when the giver first touches the object, $t_{OGC}$, and the instance when object transfer occurs, $t_{RGC}$. These instances are found by identifying the point in time when the distance between giver's hand and object reaches a minimum, and when the distance between giver's hand and receiver's hand reaches a minimum, respectively. Fig. \ref{fig:prelim_search} shows an example from a typical trial.
The end of the reaching motion, $t_{end}$, is then simply $t_{RGC}$. 

\begin{figure}[t]
    \vspace{-5mm}
     \centering
     \begin{subfigure}[b]{0.23\textwidth}
         \centering
         \includegraphics[width=\textwidth]{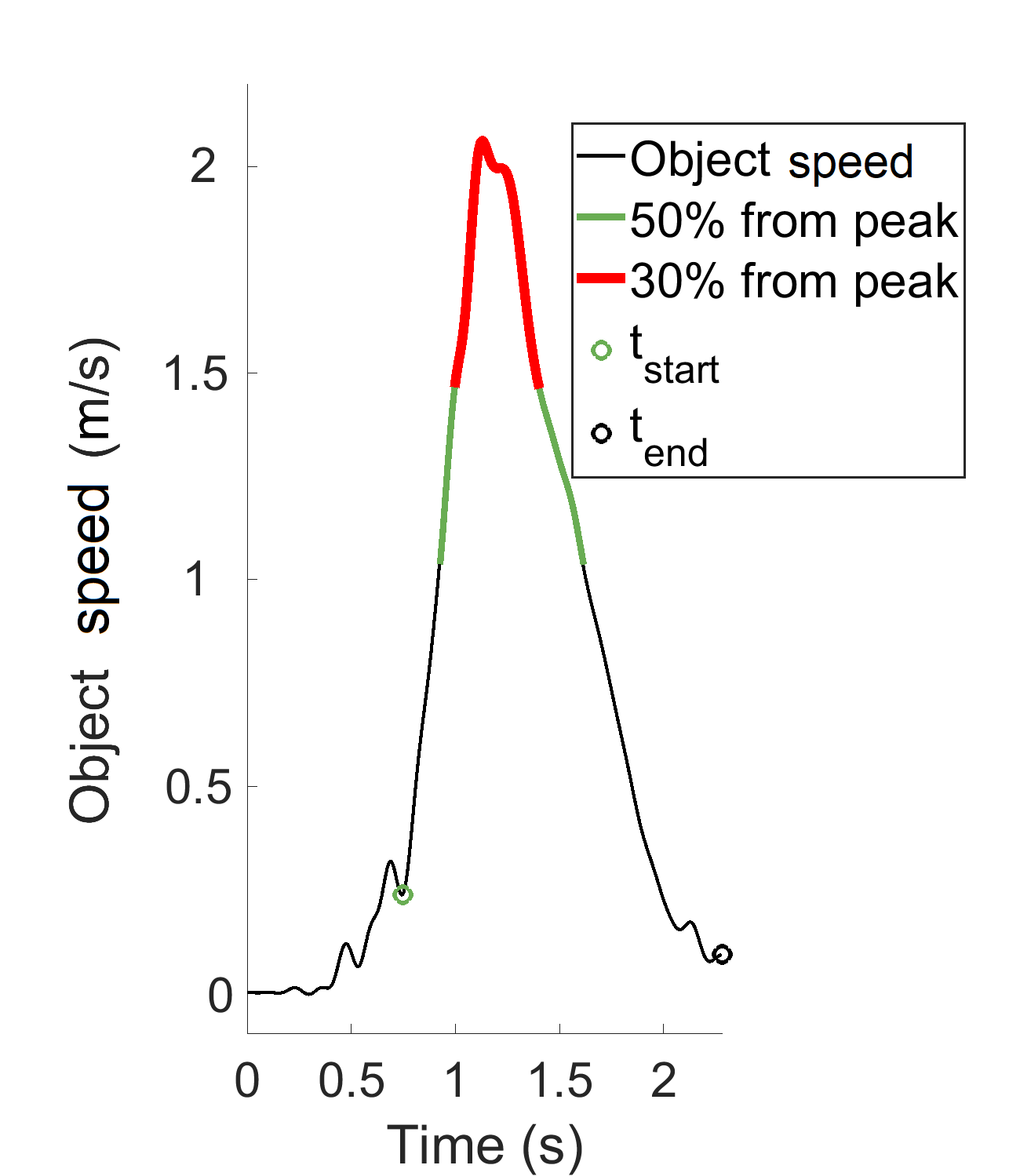}
        \caption{} \label{fig:tendency_x}
     \end{subfigure}
     \hfill
     \begin{subfigure}[b]{0.23\textwidth}
         \centering
         \includegraphics[width=\textwidth]{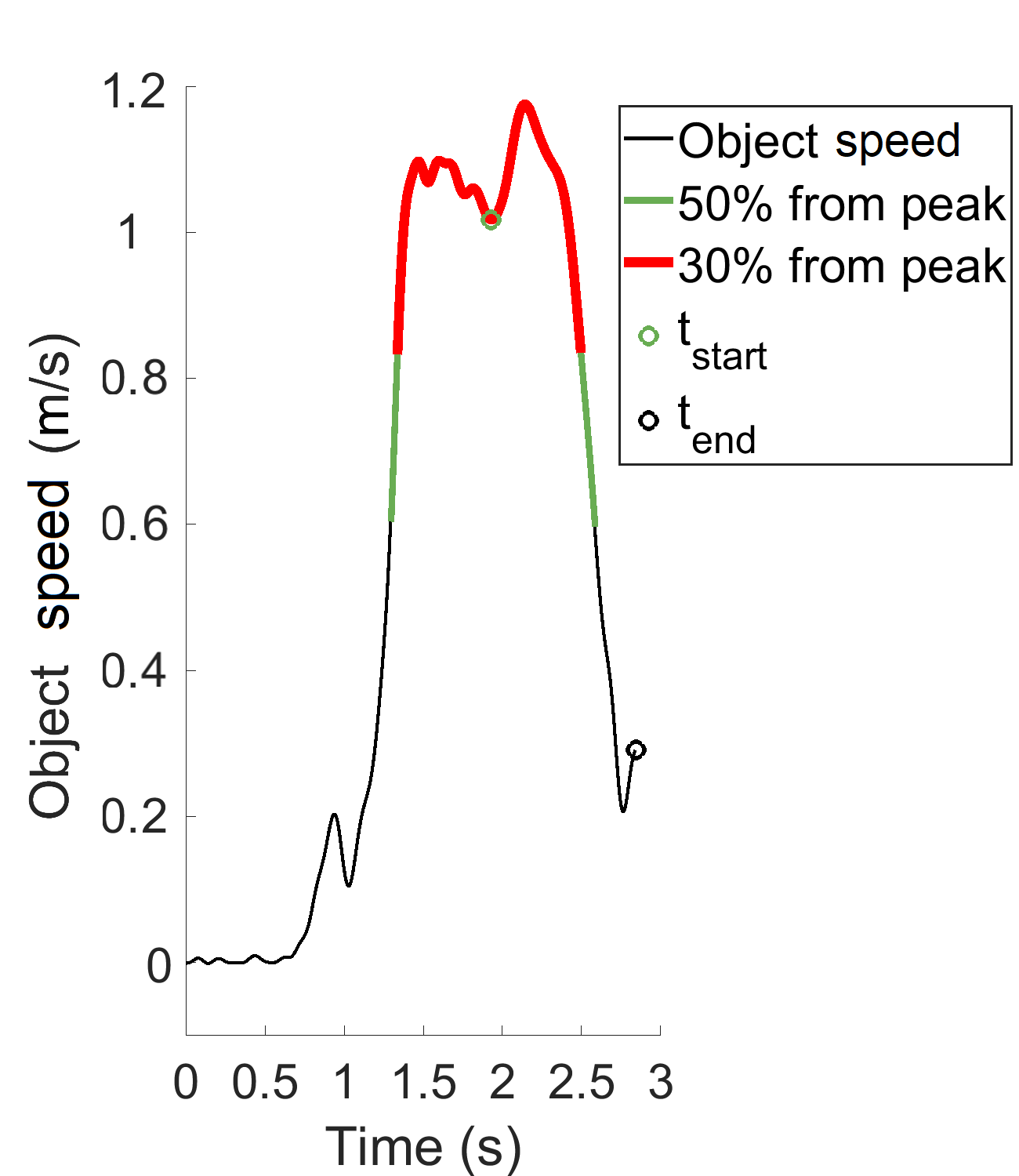}
        \caption{} \label{fig:tendency_y}
     \end{subfigure}
        \caption{Typical speed profile of an object. (a) Case 1, Giver immediately hands an object over after picking up. (b) Case 2, Giver brings the object closer to themselves first after picking up the object before handing over.}
        \label{fig:object_vel}
    \end{figure}

Determining the start of the handover reaching motion, $t_{start}$, however, requires an extra step. Inspecting the data, we observed two common giver tendencies: Case 1, reaching directly towards receiver after object pick up ($73.97\%$ of trials), and Case 2, first bringing the object closer to themselves before reaching towards receiver ($26.03\%$ of trials). Fig. \ref{fig:object_vel} shows example speed profiles demonstrating the two tendencies. In the latter case, the giver slows down slightly as they bring the object closer to themselves, manifesting a trough (confirmed through inspection of data playback using Vicon software). We detect this by inspecting if the region where the object's speed is $>$70\% peak speed (Fig. \ref{fig:object_vel} red portion) contains any local minima. In Case 2, the start of the handover reaching motion, $t_{start}$, is determined to be the local minimum, while in Case 1, $t_{start}$ is determined to be the last local minimum that is $<$50\% peak speed (Fig. \ref{fig:object_vel} black portion).   

\begin{figure}[t]
     \centering
     \begin{subfigure}[b]{0.23\textwidth}
         \centering
         \includegraphics[width=\textwidth]{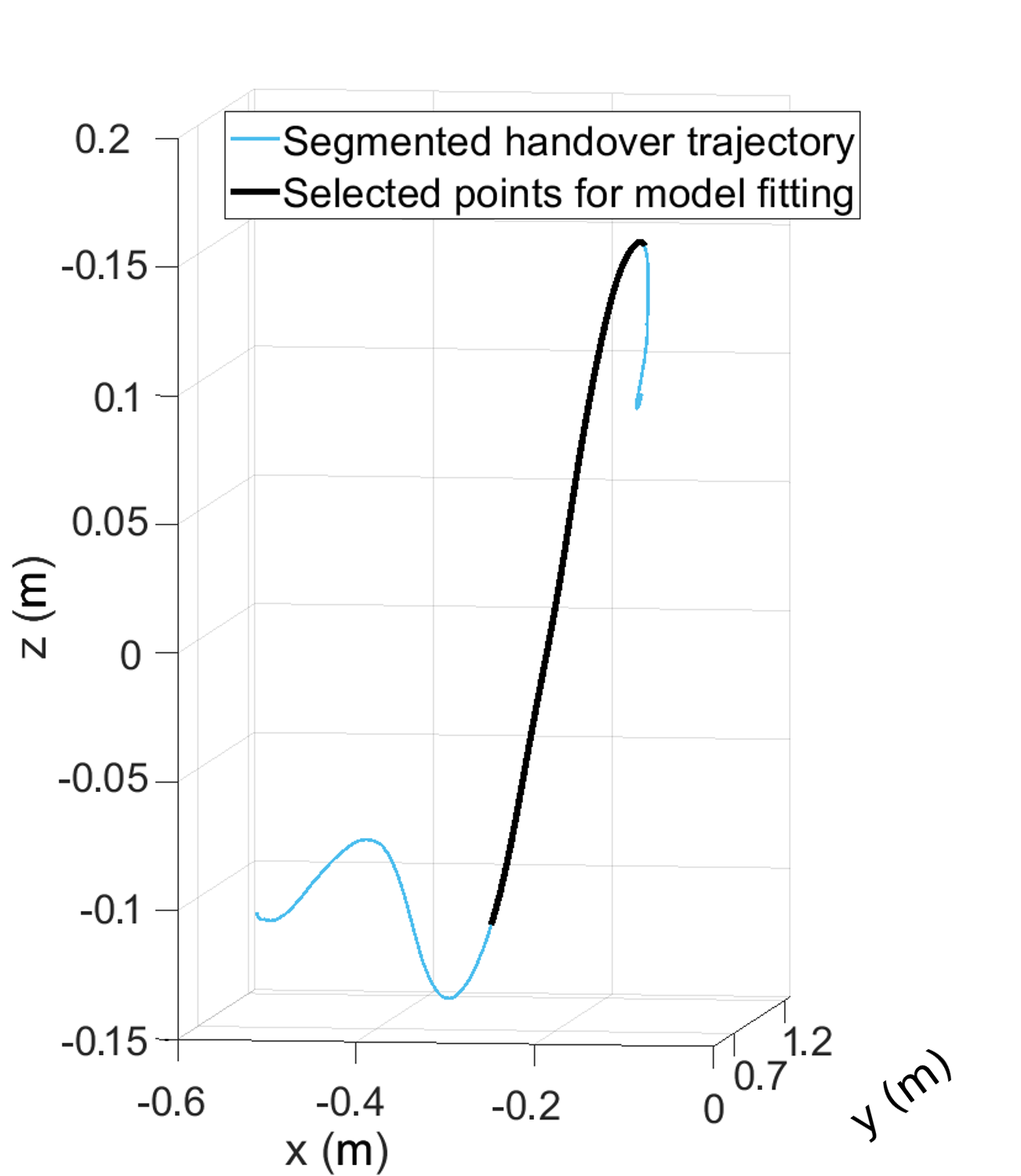}
        \caption{} 
        \label{fig:3D_segmentation}
     \end{subfigure}
     \hfill
     \begin{subfigure}[b]{0.23\textwidth}
         \centering
         \includegraphics[width=\textwidth]{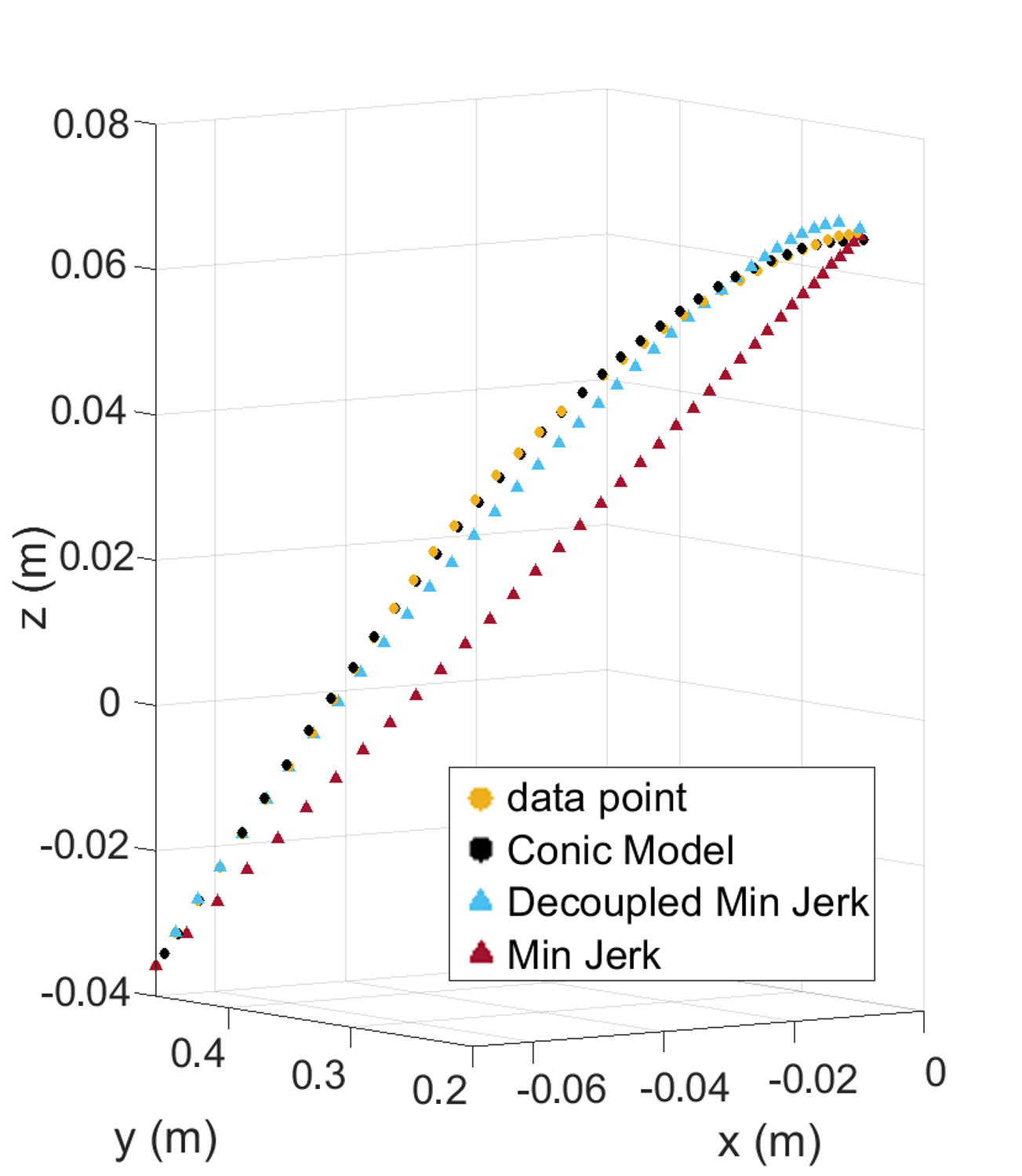}
        \caption{} \label{fig:tendency_y}
     \end{subfigure}
        \caption{Handover trajectory example from a typical trial. (a) Segmented handover reaching trajectory. (b) Example handover trajectory and the fitted models.}
        \label{fig:3D_fit}
    \end{figure}

To filter out noisy data points sometimes found near beginning and end of the segmented handover reaching motion, when fitting each trajectory model, we included only points that lie within 3 standard deviations of the best fit plane, estimated using $\Delta$\% of the segment centered around the first earlier point that is $\delta$ mm below the vertical peak of the segment. We empirically set $\Delta$=20 and $\delta$=5  to capture as many coplanar points within the handover trajectory as possible, while excluding data points belonging to arm motion prior to and after the handover trajectory. Fig. \ref{fig:3D_segmentation} shows the segmented handover trajectory and data points included for model fitting from one typical example trial.
    
\subsection{Error Calculation}
For each fitted model, we compute the fitting error as the average Euclidean distance from each data point to the correspondent point on the fitted model:
\begin{equation}
    err = \frac{1}{N} \sum^N_i ||\boldsymbol{r}_{gi} - \boldsymbol{r}_i||
\label{eq:error}
\end{equation}
where $\boldsymbol{r}_i$ is the $i^{th}$ data point in reaching trajectory, and
$\boldsymbol{r}_{gi}$ is the correspondent point on the fitted model. We define the correspondent point as the point on the fitted curve with the smallest Euclidean distant (closest point) from the data point. For each motion model, the correspondent point $\boldsymbol{r}_{gi}$ to each data point $\boldsymbol{r}_i$ is computed as follows:

\begin{itemize}
    \item Minimum Jerk model: Since the Minimum Jerk trajectory path is defined as a straight line in 3D, finding the correspondent point reduces to finding the closest point on a line to a given point. This can be easily solved analytically. Let $\boldsymbol{P}_1$ be the start point, and $\boldsymbol{P}_2$ be the final point, on the reaching trajectory. The correspondent point is given by
    \begin{equation}
        \boldsymbol{r}_{gi} = \boldsymbol{P}_1 - (\boldsymbol{P}_2 - \boldsymbol{P}_1)\left(\frac{(\boldsymbol{P}_1 - \boldsymbol{r}_i)\cdot(\boldsymbol{P}_2 - \boldsymbol{P}_1)}{|\boldsymbol{P}_2 - \boldsymbol{P}_1|^2}\right)
    \end{equation}
    \item Decoupled Minimum Jerk and Elliptical (Conic) model: Both models yield reaching motions that are planar. We first fit a plane $\mathit{P}$ to the segmented data points, then apply a rotation $\mathcal{R}$ to the reference frame to make the plane $\mathit{P}$ horizontal. The z component of the data points are then removed to collapse all points onto $\mathit{P}$. We first find the closest point in the plane $\mathit{P}$ through optimization. Then, we rotate the point using $\mathcal{R}^{-1}$ to find the correspondent point $\boldsymbol{r}_{gi}$ in 3D.
\end{itemize}

\subsection{Comparison of Motion Models}
To determine which model best fits unconstrained handover reaching motions, we conducted an ANOVA analysis, followed by t-tests with Bonferroni correction (alpha = 0.05) to compare the fitting error of the three models (Conic, Minimum Jerk, Decoupled Minimum Jerk). We then examined the conic fit results to determine the percentage of elliptical, parabolic, and hyperbolic curves.

\section{RESULT}
        \label{sec:results}
\begin{figure}[t]
    \centering
    \includegraphics[width=0.40\textwidth]{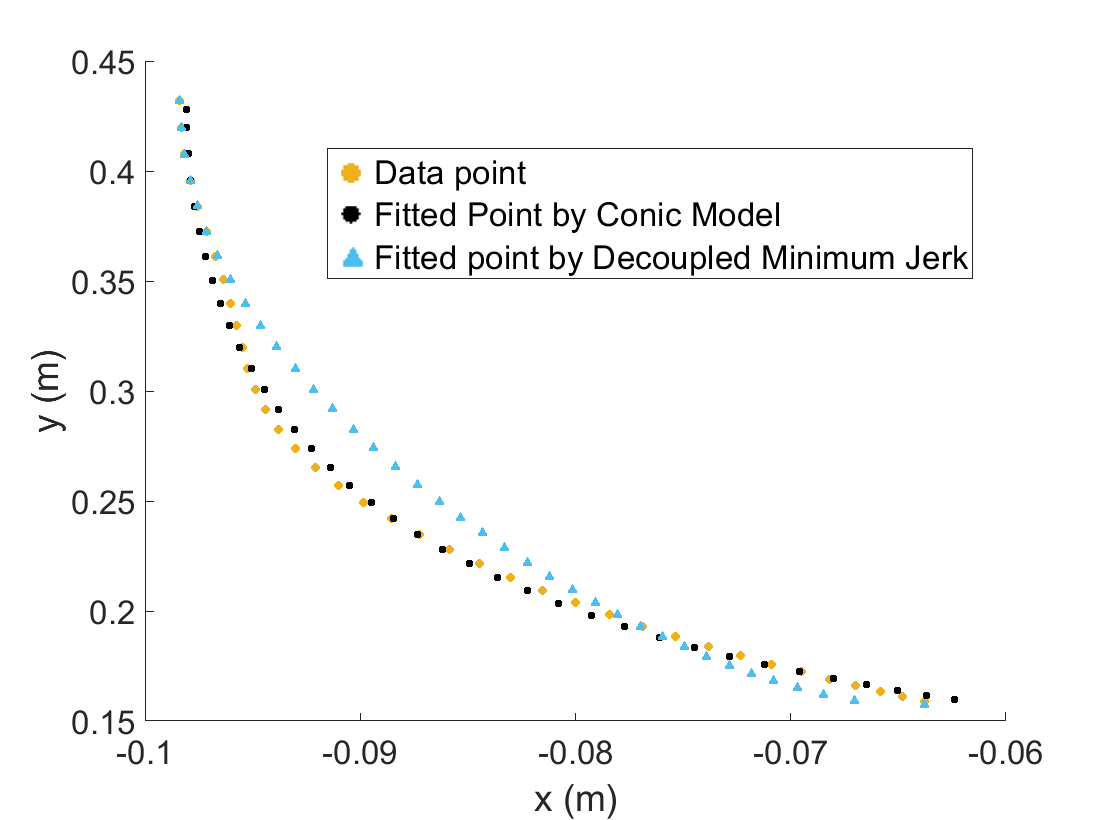}
    \caption{Fitted Conic and Decoupled Minimum Jerk models projected onto the best fit 2D plane from an example trial. Plots rotated to align conic major and minor axis with x and y axis.}
    \label{fig:2D_fit}
    \end{figure}

        Out of the 1200 handovers in the dataset, we excluded 4 trials with missing trajectory data and 1 outlier where participants handover with their left hand.
Fig. \ref{fig:3D_fit}b shows the three fitted models for an example trial in 3D space. Fig. ~\ref{fig:2D_fit} shows the fitted Conic and Decoupled Minimum Jerk models for the same example trial projected on the 2D best fit plane.
Table \ref{tab:results} summarizes resulting fitting error for each model. Examining individual trials, among the 1195 fitted handover motions, there are 752 (62.9\%) samples where conic model fits best, 442 (37.0\%) where Decoupled Minimum Jerk model fits best, and 1 (0.08\%) where Minimum Jerk trajectory fits best. When conic model is the best fit, the average fitting error for the Decoupled Minimum Jerk model is $7.2\pm10.0$ times larger. However, when the Decoupled Minimum Jerk model is the best fit, the average fitting error for the Conic model is only $2.9\pm3.6$ times larger. 

\begin{table}[b]
\caption{Model fitting error mean and standard deviation.}
\label{tab:results}
\begin{tabular}{|l|c|c|c|}
\hline
                   & \multicolumn{1}{l|}{Conic} & \multicolumn{1}{l|}{Decoupled Min Jerk} & \multicolumn{1}{l|}{Min Jerk} \\ \hline
Mean (mm)        & 2.0                       & 2.4                                        & 14.3                             \\ \hline
Std. Deviation (mm) & 3.2                       & 3.9                                        & 12.4                             \\ \hline
Max (mm) & 34.2                       & 34.5                                        & 121.9                             \\ 
\hline
\end{tabular}
\end{table}

Result from ANOVA comparing human handover reaching motion fitting error among the three models showed that there are significant differences ($F(2,358)=970.7, p<0.0005$). Post hoc t-test revealed that both Decoupled Minimum Jerk and Conic model fitted unconstrained handover reaching motions better than Minimum Jerk trajectory ($t(1194)=39.9,p<0.0005;t(1194)=33.7,p<0.0005$). Furthermore, Conic model fitted better than Decoupled Minimum Jerk trajectory ($t(1194)=3.03,p=0.007$). Statistical analysis showed that Conic model is the best fit model. Examining the fitted Conic model, 407 (34.1\%) samples were elliptical, 705 (59.0\%) hyperbolic, and 83 (6.9\%) parabolic.

\section{DISCUSSION}
        \label{sec:discussion}
\begin{figure}[t]
    \centering
    \includegraphics[width=.40\textwidth]{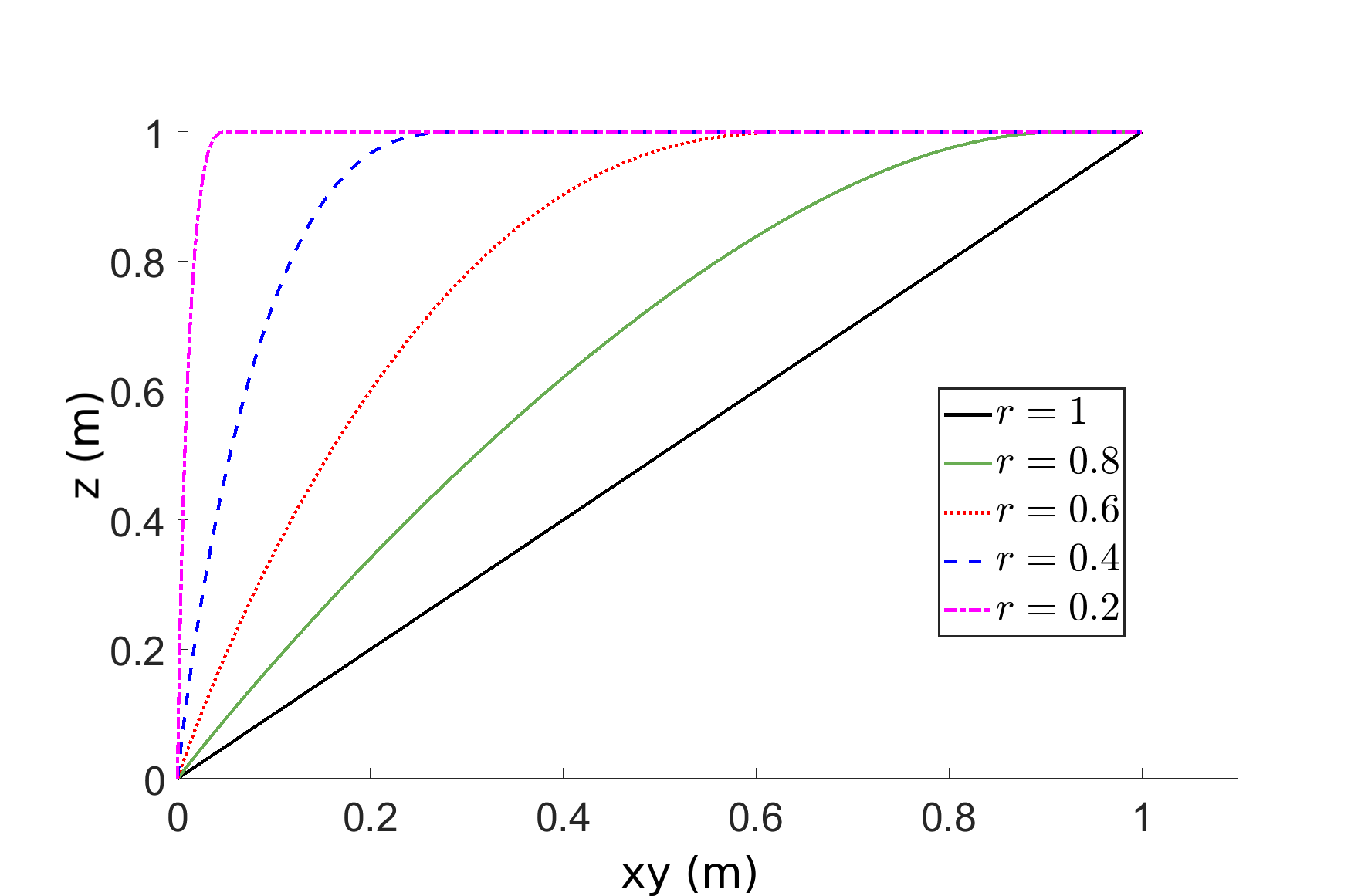}
    \caption{Resulting Decoupled Minimum Jerk Trajectories for different duration ratios $r=\frac{t_z}{t_xy}$. Note that $r = 1$ yields a Minimum Jerk Trajectory}
    \label{fig:DMJn0_config}
    \end{figure}

        The original Minimum Jerk trajectory model was formulated for reaching motions and follows a straight line trajectory for point-to-point motions. This was found to be contrary to natural human handover reaching motions, which were observed to be curved. The Decoupled Minimum Jerk model was subsequently proposed to allowed a curved path, as inspecting the velocity profile along different axes of human reaching motions, it was found that the z-axis motion tends to terminate sooner than that in the xy plane, prompting the decoupling of the axes and setting different motion durations \cite{Huber2009}. Consequently, the relationship between the motions along each axis are non-linear, creating a curve in space instead of a straight line. Although this approach is more similar to natural human motions, the residual at the end of the Decoupled Minimum Jerk trajectory yields an unnatural straight line at the end of the motion. For example, without loss of generality, let $t_z < t_{xy}$. At $t_z$, the motion along z-axis has finished while motion in the xy plane has not. After this, motion along z-axis simply ceases, while the reaching motion model continues to move in xy plane in a straight line. Fig. \ref{fig:DMJn0_config} shows the effects of $t_z$ to $t_{xy}$ ratio on the shape of the curve. The Conic model, on the other hand, does not suffer from this phenomenon. Hence, of the three models considered, the Conic model is the only one that yields curved motion along the entire reaching trajectory. 

We have chosen the Minimum Jerk Model as the baseline since it is a well-established model used for handover trajectories. Although the conic model's better fit may be attributed to having more parameters, and one may be tempted to simply use an even higher order function to achieve a better fit, this would increase the computational cost for online trajectory generation and risk overfitting. We have confirmed by Bayesian Evidence (following \cite{Sheikholeslami2018}) that indeed, a second order polynomial (conic) best fits the dataset. 

\subsection{From Path to Trajectory}
Path specifies positional aspect of a motion while trajectory specifies positional and temporal aspects of a motion. The Elliptical (Conic) model \cite{Sheikholeslami2018} is a path model while the Minimum Jerk \cite{Hogan1982} and Decoupled Minimum Jerk \cite{Huber2009} models are trajectory models. Note that while $\theta$ in Eq. \ref{eq:ellipse_parametric} can be expressed in terms of time to obtain a trajectory for the Conic model, previous study  \cite{Sheikholeslami2018} had not examined the temporal aspect and considered only the path. In our study we also compared the path (without temporal aspect) of each motion model. Although the path generated by a Decoupled Minimum Jerk trajectory was found to fit the reaching motion well, we found that the error increases significantly if we use the actual trajectory and take the model's temporal aspect into consideration (i.e., defining point correspondence for Eq. \ref{eq:error} using time rather than closest point based on spatial distance). Hence, by finding an appropriate time parameterization of $\theta$ for Eq. \ref{eq:ellipse_parametric}, we can potentially construct a Conic trajectory model that better fits human handover reaching motion as well.

\subsection{Unconstrained Handovers}
Unlike most existing studies on reaching motions, which focused on constrained (seated, tabletop) tasks, we have examined unconstrained handovers, as we are interested in enabling handovers with people for humanoids and robots that operate in natural everyday settings. This meant that participants were no longer restricted to using only arm motion, but could utilize their whole body. We discovered that, contrary to literature reporting that human reaching motion in constrained tasks is elliptical \cite{Sheikholeslami2018}, a majority of human reaching motion in unconstrained handovers are hyperbolic instead. We observed that unconstrained participants naturally utilize full body motion to carry out handovers, with a combination of the following body motions used (in addition to arm motion).

\textit{Torso Twisting}: Objects were placed beside or behind the giver. As a result, the giver naturally needs to turn their body to pick up the objects. However, torso twisting was observed to take place not only during the pick up phase, but through out the entire handover task. Participants used torso twisting to bring the object from the side/back to the front (towards the receiver), as well as to extend their arm's reach when delivering the object to the receiver. Hence, the hand trajectory incorporates torso rotational motion and shoulder translational motion.

\textit{Leaning}:
Towards the end of the handover, the giver sometimes leans towards the receiver to deliver the object. This leaning motion can potentially be pre-planned, or if the giver misjudged the object transfer location, after the arm has been fully reached out, and torso twisting utilized, the giver needs to lean further to cover the remaining gap. 

\textit{Stepping}: Givers occasionally take steps towards the receiver during handover. This potentially adds a translational component
to the hand trajectory with respect to the world frame. Consequently, the end of the hand trajectory may be elongate and flatten. However, literature shows that givers tend to begin their reach only after they are sufficiently close to the receiver \cite{Basili2009}, and in our analysis, we found that most givers took zero or only one step. Hence, the observed trajectories in our studies are mostly a combined result of the other aforementioned motions.

\subsection{Rare Handover Cases}
We observed some unexpected/outlier behaviour in the dataset. For example, some givers only picked up the object and held the object near their torso, without much "reaching" motion towards the receiver, and waited for the receiver to reach out to take the object. Another example is that sometimes the giver has to "hand down" the object because the receiver did not reach out their hand sufficiently to meet the giver at the midpoint between them. These interesting outliers may provide insights on human behaviour and how we should program human-robot handovers for humanoids operating in the real world. 

\subsection{Towards Fluent Human-Robot Handovers}
Humanoid robots will be expected to perform frequent handovers with users in many applications. Hence, their ability to perform this task well is crucial.
Existing studies tend to argue the use of a Minimum Jerk or Decoupled Minimum Jerk trajectory for handover reaching motions by showing that their velocity profiles look similar to observed human reaching motions \cite{Shibata1997,Huber2008,Huber2009}. They show that when implemented onto robot handover reaching motions, it produces more positive subjective perception of the robot. However, they have not actually fitted the proposed trajectory models to human reaching motions. We have empirically shown that the Conic model fits human handover reaching motions better than the existing Minimum Jerk and Decoupled Minimum Jerk trajectories. 

Furthermore, existing studies have been largely restricted to constrained (seated, tabletop) tasks. However, humanoids will need to perform handovers in a wide range of unconstrained situations. We have shown that human reaching motions in unconstrained handovers, while still conical, is no longer restricted to elliptical, as in constrained tasks \cite{Sheikholeslami2018}. There is also coordination of full body motion.

With the empirically verified Conic model, and the discovery that unconstrained handover motions are not restricted to elliptical motions, we can use this knowledge to implement more fluent human-robot handovers. For a robot giver, we can implement more humanlike reaching motions based on the Conic model. However, as we observed that humans use coordinated full body motion for handovers, a humanoid robot giver may also need to consider how to coordinate its full body motion, instead of planning only arm motion, as most existing handover motion planners do \cite{Shibata1997,Huber2008,Huber2009}. 
As a receiver, the robot can observe the onset of the human giver's reach, and use the Conic model to estimate the endpoint to predict the point of object transfer. This will then allow greater flexibility, and enable the robot to plan and begin executing it's reaching motion for intercepting the object early on during the reaching phase of the handover, unlike existing robots that tend to wait until the giver finishes reaching, or can only hand over the object at a predefined location \cite{Kshirsagar2020,Micelli2011,Chan2015}. As handover is a ubiquitous task for humanoids working with people, improving their competence in this fundamental task is expected to improve their overall interaction with users in a wide range of applications.


\section{CONCLUSION AND FUTURE WORK}
        \label{sec:conclusion}
        We have experimentally compared the existing Minimum Jerk, Decoupled Minimum Jerk, and Conic models for unconstrained human handover reaching motions. We showed that while the Conic models fits best, unlike seated tabletop reaching motions that were mostly elliptical, there is a split between elliptical and hyperbolic motion models. Our paper has provided the first experimental validation of reaching models for human handover reaching motions. Results suggests that the Conic model may be used to generate more humanlike motions compared to the well-established Minimum Jerk models. Furthermore, unlike solo reaching motions which are mostly elliptical, a mix of elliptical and hyperbolic motions should be expected in handovers.

While we have shown that the Conic model fits human handover reaching motions better than existing Minimum Jerk-based models, the Minimum Jerk model still has benefits. Indeed the Minimum Jerk model is a long-established model and it has been tested experimentally for human reaching motions. One potential direction for future work may be finding a unified model combining Conic and Minimum Jerk models, where the Conic model may describe the path, while the Minimum Jerk model the temporal aspect of the reaching motion. Our future work will also include implementing and testing elliptical (conic) robot handover motions in user studies.

\bibliographystyle{IEEEtran}
\bibliography{ref}

\end{document}